\documentclass[12pt]{amsart}
\usepackage{tikz}
\usepackage[utf8]{inputenc}
\usepackage{hyperref}
\usepackage[utf8]{inputenc}
\usepackage{amsmath}
\usepackage{amsfonts}
\usepackage{amssymb}
\usepackage{amsthm}
\usepackage{stmaryrd}
\usepackage{eucal}
\usepackage{graphicx}
\usepackage{indentfirst}

\usepackage{multicol}

\usepackage{mathrsfs}

\usepackage[T1]{fontenc}

\normalfont

\newtheorem{thm}{Theorem}[section]
\newtheorem{prop}[thm]{Proposition}
\newtheorem{lemma}[thm]{Lemma}
\newtheorem{corr}[thm]{Corollary}

\newtheorem{introthm}{Theorem}%
\newtheorem{introprop}[introthm]{Proposition}%
\newtheorem*{prop*}{Proposition}

\theoremstyle{definition}
\newtheorem{opr}[thm]{Definition}

\theoremstyle{remark}
\newtheorem{rem}[thm]{Remark}
\newtheorem{ntn}[thm]{Notation}

\makeatletter
\@addtoreset{equation}{section}
\makeatother

{\end{equation}%
	}%

\newcommand{\bR}{{\mathbb R}}

\newcommand{\bZ}{{\mathbb Z}}

\newcommand{\cC}{{\mathcal C}}

\renewcommand{\phi}{\varphi}

\newcommand{\Mor}{{\rm Mor} \,}

\newcommand{\bP}{\mathbb P}

\newcommand{\op}{{\sf op}}

\renewcommand{\lim}{\operatorname{\varprojlim}}

\newcommand{\KL}{\operatorname{{KL}}}
\newcommand{\AL}{\operatorname{{AL}}}

\title{A Neural Network for Semigroups}
\author{Edouard Balzin, Boris Shminke}

\begin{document}

\maketitle

\begin{abstract}
    Tasks like image reconstruction in computer vision, matrix completion in recommender systems and link prediction in graph theory, are well studied in machine learning literature. In this work, we apply a denoising autoencoder-based neural network architecture to the task of completing partial multiplication (Cayley) tables of finite semigroups. We suggest a novel loss function for that task based on the algebraic nature of the semigroup data. We also provide a software package for conducting experiments similar to those carried out in this work. Our experiments showed that with only about 10\% of the available data, it is possible to build a model capable of reconstructing a full Cayley from only half of it in about 80\% of cases.
\end{abstract}

\setcounter{tocdepth}{1}
\tableofcontents

\section{Introduction}

\subsection*{Semigroups} A semigroup structure is one of the basic algebraic structures that one can put on a set $S$. It consists of a binary operation $S \times S \to S, (x,y) \mapsto x \cdot y$ that satisfies the associativity identity: that is, for all $x,y,z$ in $S$, one requires $x\cdot (y \cdot z) = (x \cdot y) \cdot z$. The examples of such structures are many: groups and monoids, for example, come equipped with associative binary operations. For another example, given a (small) category $\cC$, denote $\Mor \cC$ the set of all morphisms, take $S = \Mor \cC \cup \{0 \}$ and put $f \cdot g = f \circ g$ if the composition is well defined and $0$ otherwise. If $\cC$ has more than one object, the result will be a semigroup that does not come from a monoid. Semigroup examples also appear from automata.  

If we restrict our attention to finite semigroups, the existing classification tables \cite{SMALLSEMI} are very counter-intuitive to any mathematician familiar with finite groups. Semigroups, classified up to an isomorphism or anti-isomorphism (Definition \ref{defequivalencesemigroup}), exist in abundance, as per Table \ref{table0} (to compare, there is only one finite group structure on a set of seven elements). While certain results of classification exist, the most known one being the Krohn–Rhodes decomposition theory \cite{PIN}, understanding semigroups of cardinality (or order) up to 20 remains a challenging task.  

\begin{table*}
\label{table0}
\caption{Number of equivalence classes of semigroups up to isomorphism or anti-isomorphism \cite{SMALLSEMI}.}
\begin{tabular}{c|c}
\hline
Cardinality & \# semigroups up to equivalence \\
\hline
1 & 1 \\
2 & 4 \\
3 & 18 \\
4 & 126 \\
5 & 1,160 \\
6 & 15,973 \\
7 & 836,021 \\
8 & 1,843,120,128 \\
9 & 52,989,400,714,478 \\
\hline
\end{tabular}
\end{table*}

\subsection{Machine learning of semigroups} This paper is our first attempt to see if one can approach questions about semigroups using the methods of machine learning. As a first question, we wanted to see how we could explain what a semigroup is to a neural network, so that the latter captures the algebraic character of the semigroup structure. 

Given a finite set $S$, we can order its elements and present any binary operation $m: S \times S \to S$ as a multiplication table. A couple of examples is given in Table \ref{tablesemigroupexamples}. Such tables can be transformed into a neural network input if viewed as tensors in $(\bR^n)^3$, where $n$ is the cardinality of $S$. Such a tensor denoted $(M_{ijk})$ takes values $M_{ijk}=1$ if the $i$'th element times the $j$'th element equals the $k$'th element, and $M_{ijk}=0$ in all other cases.

\begin{table*}
\caption{Multiplication tables of Klein 4-group $\bZ/2\bZ \times 
\bZ/2\bZ$ and of a 3-nilpotent semigroup on 5 elements.}
\label{tablesemigroupexamples}
\begin{tabular}{|c||c|c|c|c|}
\hline
 & 1 & 2 & 3 & 4  \\
\hline
\hline
1 & 1 & 2 & 3 & 4 \\
\hline
2 & 2 & 1 & 4 & 3 \\
\hline
3 & 3 & 2 & 1 & 2 \\
\hline
4 & 4 & 3 & 4 & 1 \\
\hline
\end{tabular}
\quad
\begin{tabular}{|c||c|c|c|c|c|}
\hline
 & 1 & 2 & 3 & 4 & 5  \\
\hline
\hline
1 & 1 & 1 & 1 & 1 & 1 \\
\hline
2 & 1 & 1 & 1 & 1 & 1 \\
\hline
3 & 1 & 1 & 2 & 1 & 2 \\
\hline
4 & 1 & 1 & 1 & 2 & 1 \\
\hline
5 & 1 & 1 & 2 & 1 & 2 \\
\hline
\end{tabular}
\end{table*}

The lack of classification of semigroups of higher order led us to imagine, then, the following question: can one start with a multiplication table that is partially filled, and ask a neural network provide a completion? This can be formulated as an autoencoder \cite{deeplearningbook} problem on $(\bR^n)^3$.  

In this paper, we have chosen $n=5$. As noted above, semigroups of cardinality $5$ are already classified, and while our goal would be to move into higher cardinalities, it is still of use to study a known list of semigroups to propose a kind of neural network that is useful for understanding semigroup structure. Unlike in the problem of solving sudoku \cite{sudoku} (also a problem of table completion that inspired some reflection behind this project), we cannot rely on image-processing techniques and related convolutional neural networks as they do not adequately measure the structure of semigroups. The analysis of $n=5$ case helped us to formulate the problem and design the architecture that worked. 

The $1160$ equivalence classes of $n=5$ correspond (as can be seen from \cite{SMALLSEMI, neural-semigroups}) to $183732$ different tables, which is provides a sufficient amount of data for our purposes. For the input of the autoencoder in this case, both for training and testing, we take a table that is produced from a semigroup by forgetting some multiplications, which corresponds to ``erasing'' certain cells in the multiplication table. 
The output tensor, the value of the autoencoder on such a partially filled table, would have to correspond to $a)$ a table that $b)$ is associative. In the context of supervised learning, the condition $b)$ can be enforced with different choices of loss functions. One choice would simply penalise the difference between the reconstructed table and the original one. This choice is unnatural mathematically as tables can have non-unique completions (Table \ref{tablenonunique}) and also relies on knowing the resulting semigroup. 

\begin{table*}
\caption{Different semigroups giving the same partially filled table.}
\label{tablenonunique}
\begin{tabular}{|c||c|c|c|c|c|}
\hline
 & 1 & 2 & 3 & 4 & 5  \\
\hline
\hline
1 & 1 & 1 & 1 & 1 & 1 \\
\hline
2 & 1 & 1 & 1 & 1 & 1 \\
\hline
3 & 1 & 1 & 2 & 1 & 2 \\
\hline
4 & 1 & 1 & 1 & 2 & 1 \\
\hline
5 & 1 & 1 & 2 & 1 & 2 \\
\hline
\end{tabular}
$\, \Rightarrow \,$
\begin{tabular}{|c||c|c|c|c|c|}
\hline
 & 1 & 2 & 3 & 4 & 5  \\
\hline
\hline
1 & 1 & 1 & 1 & 1 & 1 \\
\hline
2 & 1 & 1 & 1 & 1 & 1 \\
\hline
3 & 1 & 1 &   &   &   \\
\hline
4 & 1 & 1 &   &   &   \\
\hline
5 & 1 & 1 &   &   &   \\
\hline
\end{tabular}
$\, \Leftarrow \,$
\begin{tabular}{|c||c|c|c|c|c|}
\hline
 & 1 & 2 & 3 & 4 & 5  \\
\hline
\hline
1 & 1 & 1 & 1 & 1 & 1 \\
\hline
2 & 1 & 1 & 1 & 1 & 1 \\
\hline
3 & 1 & 1 & 2 & 2 & 2 \\
\hline
4 & 1 & 1 & 2 & 2 & 2 \\
\hline
5 & 1 & 1 & 2 & 2 & 2 \\
\hline
\end{tabular}
\end{table*}

Another, called the associator loss (\ref{associatorloss} below), translates the associativity of semigroup multiplication into a certain probabilistic function on the neural network. This choice accepts associative answers different from the original table and, interestingly, does not require one to actually present a semigroup that completes the partially filled table. 

\subsection{Acknowledgements} The authors of this paper are very grateful to Jordan Emme, Wesley Fussner and Carlos Simpson for their remarks. This work has been supported by the French government, through the 3IA C\^ote d’Azur Investments in the Future project managed by the National Research Agency (ANR) with the reference number ANR-19-P3IA-0002.

\section{Basics of semigroups}

\begin{opr}
A \emph{semigroup} is a set $S$ together with a binary operation $\cdot: S \times S \to S$, $(a,b) \mapsto a \cdot b$, that is associative: for all $a, b, c$ in $S$, one has $a\cdot (b \cdot c) = (a \cdot b) \cdot c$.
\end{opr}

One could thus say that a semigroup is an associative magma, or a non-unital monoid. Various monoids and groups provide examples of semigroups if we forget the extra properties (existence of units, inverses). In this paper, we only consider the case when the set $S$ is finite. We shall write $(S,\cdot)$ to denote a semigroup or simply $S$ when this does not lead to confusion.

\begin{ntn}
Let $(S, \cdot)$ be a semigroup. Its \emph{opposite} semigroup will be denoted $S^\op=(S,\cdot^\op)$. Its underlying set is $S$ and the multiplication operation $\cdot^\op$ is defined as $a \, \cdot^\op \, b := b \cdot a$. 
\end{ntn}

\begin{opr}
\label{defequivalencesemigroup}
A homomorphism, or simply a morphism of semigroups $(S, \cdot) \to (T, *)$ is a map of sets $f: S \to T$ such that for all $a,b \in S$, one has $f(a) * f(b) = f(a \cdot b)$. Two semigroups $S,T$ are called \emph{equivalent} if there is an isomorphism between $S$ and $T$ or between $S$ and $T^\op$ (the latter meaning that $S$ and $T$ are anti-isomorphic). 
\end{opr}
A semigroup map $f$ is an isomorphism if and only if it is a bijection of the underlying sets.

Just like with finite groups, the multiplication of finite semigroups can be described in terms of a multiplication table. Our work deals with those tables viewed as certain tensors that can be obtained as follows.

\begin{opr}
Let $(S, \cdot)$ be a semigroup and $k$ a field (one can assume $k= \bR$ for the purposes of machine learning). Its associated \emph{semigroup algebra} $k[S]$ is a non-unital $k$-algebra defined as follows. As a vector space, $k[S]=\cong k^{|S|}$ is the free vector space on $S$, and the multiplication operation is extended from $\cdot$ by $k$-bilinearity: $(\sum_i \lambda_i a_i)(\sum_j \mu_j b_j) := \sum_{i,j} \lambda_i \mu_j (a_i \cdot b_j)$.   
\end{opr}

Any semigroup morphism $S \to T$ induces a non-unital algebra morphism $k[S] \to k[T]$, and one has $k[S^\op] = k[S]^\op$.

The multiplication operation, being bilinear, can be viewed as a map $m:k[S] \otimes k[S] \to k[S]$. When $S$ is finite, the semigroup algebra $k[S]$ is finite-dimensional as a vector space, and so we can view the multiplication as a tensor $m \in (k[S]^*)^{\otimes 2} \otimes k[S]$.

From now on, assume that $S$ is finite of cardinality $n$. If one denotes $S = \{e_i \}_{i=1}^n$, then the $e_i$ and their dual linear forms $f_j$ form bases of $k[S]$ and $k[S]^*$, respectively. The multiplication $m$ can then be expressed as 
$$
m =\sum_{i,j,k=1}^{n} M_{ijk} \, e_k \otimes f_i \otimes f_j,
$$
where $M_{ijk}$ is equal to $1$ if $e_i \cdot e_j = e_k$ and is $0$ otherwise.

\begin{opr} For a finite semigroup $S$, the coefficients $(M_{ijk})$ are called the \emph{structure constants} of $S$. In this work, we shall also refer to $(M_{ijk})$ as the \emph{Cayley table} of $S$.  \end{opr}

\begin{rem} 
\label{remalgstructureconst}
If instead of $k[S]$ we considered a general non-unital $k$-algebra $A$ that is finite dimensional as a $k$-vector space, with a basis that one can also denote $e_i$, the coefficients $M_{ijk}$ obtainable in the same way from the multiplication $A \otimes A \to A$ would not consist only of $0$'s and $1$'s. The associativity condition for the basis vectors, $(e_i \cdot e_j) \cdot e_k = e_i \cdot (e_j \cdot e_k)$ gives the following equation on the coefficients:
\begin{equation}
\label{structureconstanteq}    
\sum_{m} \left(  M_{ijm} M_{mkl} - M_{iml}M_{jkm} \right) = 0.
\end{equation}
\end{rem}

A typical presentation of $M_{ijk}$ is usually done in a table format, by putting $e_k$ or simply its index $k$ in the ${ij}$-cell (as done in Table \ref{tablesemigroupexamples} above). One can imagine such a multiplication table being partially filled, with some multiplications $e_i \cdot e_j$ not specified, an idea that we formalise below.

\section{Experimental setup}
\subsection{Data representation}
Given a set $S= \{e_i \}_{i=1}^n$, consider a function $F: S \times S \times S \to [0,1]$. We would like to treat this function as a probability distribution for the potential multiplication: $\bP(e_i \cdot e_j =e_k)= F (e_i,e_j,e_k)$. For this to make sense, the function $F$ must satisfy the following condition: $\sum_k F(e_i,e_j,e_k)=1$ for all possible choices of $i$ and $j$.

\begin{opr} Call such a function $F$ a \emph{partial Cayley table}. A partial Cayley table $F$ is \emph{filled at} $1 \leq i,j \leq n$ if there exists $k$ such that $F(i,j,k):=F(e_i,e_j,e_k)=1$.
\end{opr}

Any semigroup structure on $S$ provides us with the partial Cayley table $F(i,j,k) = M_{ijk}$ that is actually filled at all $i,j$. Our definition does not guarantee however that a partial Cayley table that is filled at all $i,j$ corresponds to an associative multiplication. For this reason, define:

\begin{opr}
A partial Cayley table $F:S \times S \times S \to [0,1]$ is \emph{solvable} if there exists a semigroup structure on $S$, with structure constants $(M_{ijk})$, such that $M_{ijk} = F(i,j,k)$ for all $i,j$ at which $F$ is filled. 
\end{opr}

As noted in the introduction, a solvable $F$ can have multiple semigroup solutions. 

We can store $F\left(i,j,k\right)$ as a tensor of one axis of dimension $n^3$ (in \cite{pytorch} parlance), e.g. using a lexicographical order of triples of indices $\left(i,j,k\right)$.
These tensors are used as the main method of data representation in this work.

\begin{figure}

\includegraphics[width=0.92\textwidth]{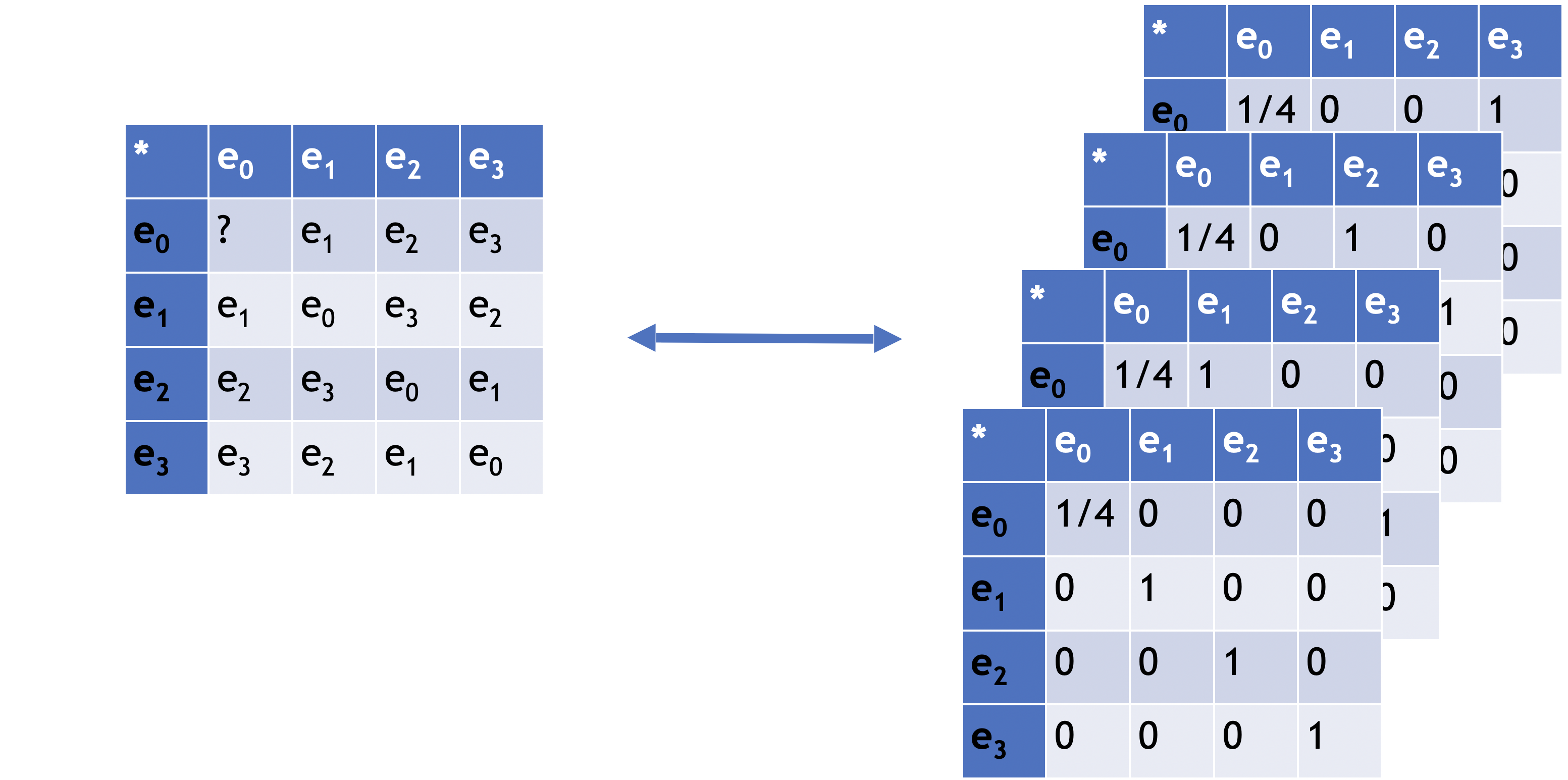}

$\,$ 
 
    \caption{Translating a multiplication table into as a partially filled $F: \{0,1,2,3 \}^3 \to [0,1]$.}
    \label{tablepartiallyfilledexample}
\end{figure}

Assume now that for a set $S$ we specified only some multiplications for a semigroup structure. This allows us to partially define the function $F$. If the result of multiplication $e_i \cdot e_j$ is not specified, we can extend by employing a uniform distribution in such cases, i.e. assume in that case that $F(i,j,k)=\frac1n$ for all $k$ (Figure \ref{tablepartiallyfilledexample}).  

\subsection{Network architecture}

One can consider a partial Cayley table as a result of distortion of a corresponding fully filled table. 
Similarly, arbitrary probabilistic tensors can be viewed as noisy counterparts of zero-or-one tensors. A well-known way to get rid of distortions and restore the original of an image is by using a denoising autoencoder \cite{deeplearningbook}.

\begin{figure}
\includegraphics[width=0.64\textwidth]{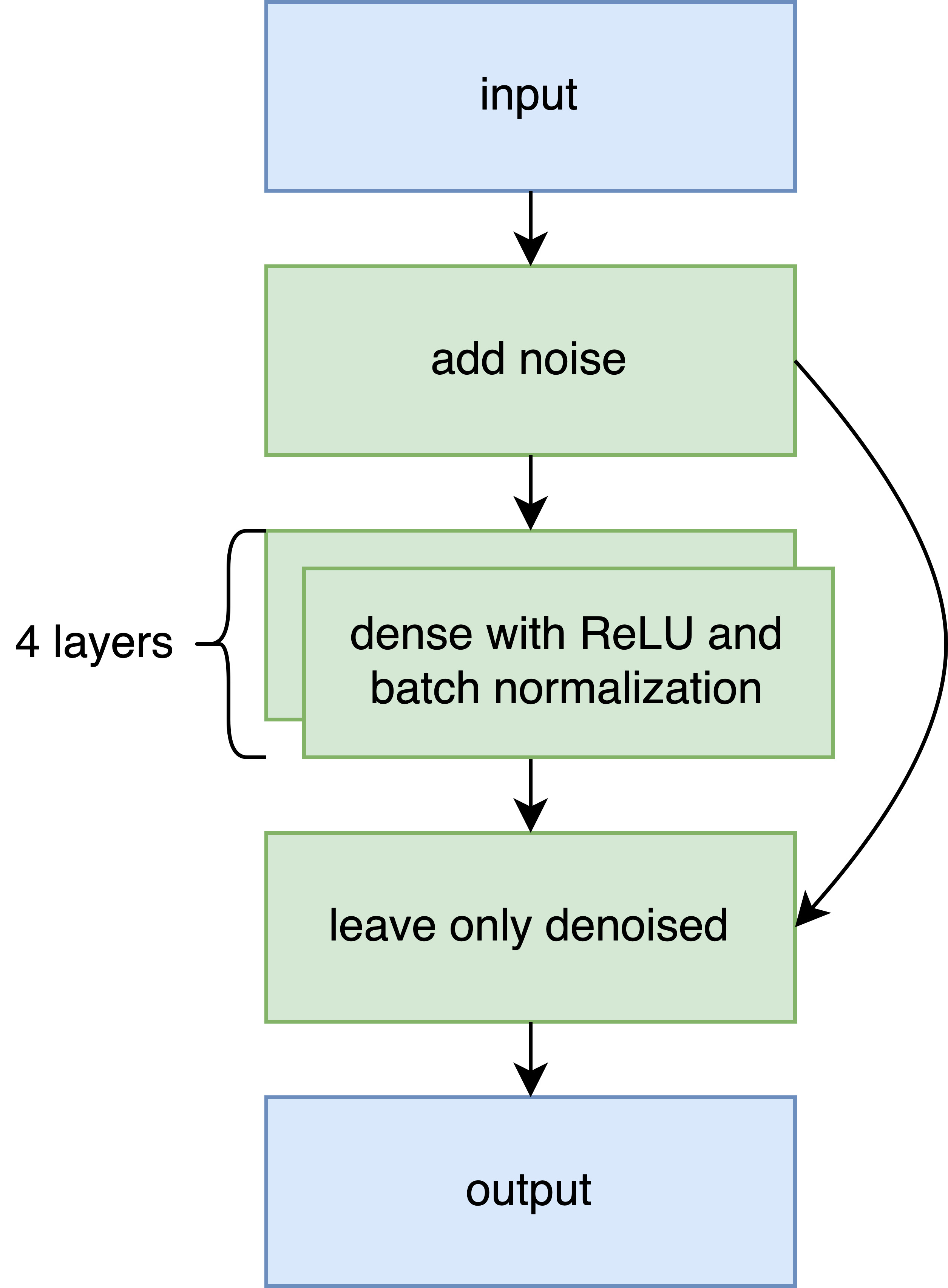}
\caption{Autoencoder architecture used to generate Cayley tables. The arrow from the input with added noise to the ``leave only denoised'' layer corresponds to restoring the values of initially known cells.}
\label{architecture}
\end{figure}

For a scheme of the autoencoder architecture used in this paper see Figure \ref{architecture}. Besides adding noise to its input, this network also cleans its output of guesses of the cells which were not masked during noise addition; these cells correspond to known fillings of the Cayley table. In other words, if the input was filled at $i,j$ so that $e_i \cdot e_j = e_l$ and the output $F\left(i,j,k\right)$ during the forward pass is a float between $0$ and $1$, it then redefined as $F\left(i,j,k\right):=0$ or $1$ corresponding to $k \neq l$ or $k=l$.

Another particular thing to note is that usually we have encoders which move from higher dimensions to lower ones. Here we have input and output both of dimension $n^3$ and the hidden layers all of dimension $n^5$.

For more details we invite the reader to consult the source code available at \cite{neural-semigroups}.
\subsection{Loss functions}
If $x$ is an input for an autoencoder and $y$ is its output, we can define its loss function $L\left(x,y\right)$ in a variety of ways.
Since in our case values of $x$ and $y$ are probabilities of a joint distributions, it could be a good idea to use some measure of dissimilarity between these two distributions, e.g. their Kullback-Leibler divergence:
$$\KL\left(x,y\right):=\sum\limits_{i=1}^nx_i\log\frac{x_i}{y_i}.$$
Note that this choice of a loss function does not explicitly enforce any notion of associativity. The problem with this function is that after applying corruption to $x$ it can often be recovered as $y$ non-uniquely, yet the loss function $(x,y) \mapsto \KL(x,y)$ will prefer $y=x$ to any other value of $y$, even if that value is associative. 

Another choice of a loss function is what we call the associator loss. First, remember that $y$ corresponds to the probability distribution $y_{ijk}=\bP\left\{e_i\cdot e_j=e_k\right\}$. Then we can calculate probabilities of double multiplications:
\begin{align}
\bP\left\{\left(e_i\cdot e_j\right)\cdot e_k=e_l\right\} &=
\sum\limits_{m=1}^n \bP\left\{e_m\cdot e_k=e_l\vert e_i\cdot e_j=e_m\right\}\bP\left\{e_i\cdot e_j=e_m\right\} \nonumber \\
&= \sum\limits_{m=1}^ny_{mkl}y_{ijm}. \nonumber
\end{align}

Now we can define the loss function as a KL-divergence between the distributions $\bP\left\{\left(e_i\cdot e_j\right)\cdot e_k=e_l\right\}$ and $\bP\left\{e_i\cdot \left(e_j\cdot e_k\right)=e_l\right\}$:
\begin{equation}
\label{associatorloss}
\AL\left(x,y\right):=\KL\left(\sum\limits_{m=1}^n y_{ijm} y_{mkl},\sum\limits_{m=1}^ny_{iml}y_{jkm}\right).
\end{equation}

This loss does not depend on $x$ but only on probabilistic associativity of $y$, and indeed it corresponds to interpreting in probabilistic terms the coefficient equation (\ref{structureconstanteq}) of Remark \ref{remalgstructureconst}.
\subsection{Noise}
In our case, the noise which autoencoder is treating corresponds to the absence of some number of cells in a Cayley table. In our experiments, both for training and testing, we take tables of semigroups of cardinality $5$. Given any table $F:S \times S \times S \to \bR$ we then add noise by re-setting $F(i,j,k)=\frac1 5$ for $i,j$ corresponding to randomly chosen $50\%$ of cells of the original Cayley table.
\subsection{Training and testing datasets}
For this work we use an extensive database of finite semigroups up to eight elements from \cite{SMALLSEMI}.
If we fix the number of elements that is less or equal to $8$, we get the number of classes of equivalence of semigroups (Definition \ref{defequivalencesemigroup}) as well as their presentation in \cite{SMALLSEMI}.
In this paper, we used semigroups of $5$ elements for experiments: as mentioned, this corresponds to $1160$ classes and $183732$ potential tables.

In detail, we proceeded by dividing this set of $1160$ equivalence classes into three subsets: training, validation, and testing in proportion $10 / 10 / 80$.
We then produced all Cayley tables of isomorphic and anti-isomorphic semigroups corresponding to these classes of equivalence, a procedure that one can view as a form of data augmentation.
Finally, we applied the noise as described in the previous section, but only to validation and testing sets. The training set gets its noise during the training process, and $50\%$ of cells to be masked are chosen at random for every batch, and are not fixed in advance for all the training process. Note that all partial tables appearing here are solvable.

\subsection{Quality metrics}
Since we train autoencoders, it is natural to use the following metrics:
\begin{opr}
The \emph{guess rate} is the percentage of outputs of a network which coincide with their inputs before applying noise.
The \emph{associative rate} is the percentage of outputs of a network which satisfy the assoiciativity condition.
\end{opr}
The associative rate appears to be a better quality metric, since we are interested not in exact reconstruction of inputs but in generating associative tables. A half-filled table can be completed to different semigroups, but the guess rate will only accept the original table for its score.

\subsection{Training process}
We trained all the networks using the Pytorch \cite{pytorch} framework, using an Adam optimizer \cite{ADAM} with the learning rate set to $0.0001$. The training was done for a maximum of $1000$ epochs with an early stopping applied if the loss did not go down for ten consecutive epochs. The training was done on \cite{colab} cloud resources and took several hours in total. We performed batch normalisation on each layer and used random network parameter initialisation.

\section{Results and discussion}
First, we note that teaching an autoencoder to simply reconstruct its input without knowing anything about associativity proved to be not only unnatural but in fact bringing poorer results. Even in terms of its main goal -- finding the original table from the input with added noise -- the KL divergence loss is less adequate than the associator loss (AL): see Table \ref{table1} for exact numbers.
One way to interpret the KL-AL guess ratio difference might be in observing that the AL network does better at the associativity task overall, in particular it does better at reconstructing the original table. 

The AL network results are rather promising. We managed to produce a full associative table given only a half of filled cells as an input in $82\%$ of cases. That is even more impressive given that we relied only on $10\%$ of all available tables from the database, thus managing to generalise to $80\%$ (which went to the test set).
Our results remain dependant on the choice of these $10\%$ tables for a training set, with deviation representing about $2$ percent of the loss (see Table \ref{table2} for the details). 

These findings make us believe that an AL-type network could also be successfully used for higher cardinalities. One could view the associator loss as the suitable "architectural adaptation" to the case of semigroups: instead of convolutional layers, we are dealing with algebraic equations written into the loss function, in probabilistic terms. Such a neural network can work on higher-dimensional data if adapted properly. In fact, it can even accept as training set partial tables that are known to be solvable; the latter can be verified with various model searchers such as Mace4/Prover9 \cite{mace4}. 

And if one is able to produce a train dataset in higher cardinalities, the trained network can be viewed as a certain generator, that produces full tables out of sets of identities corresponding to known cells in the input. We do not know if all possible tables can be produced in such a way, and verifying it for lower cardinalities is one of our future goals. For the higher cardinality, one could ask if such a neural network could produce any semigroup in its image if fed with random input. Finally, one can imagine generalisations of such neural networks to other classes of algebraic structures. 

\begin{table*}
\label{table1}
\caption{Comparison of impacts of a loss function choice. The associator loss network fares better not only at producing associative tables, but also at guessing the original table to which we applied noise.}
\begin{tabular}{l|c|c}
\hline
Loss function used & Guess rate & Associative rate \\
\hline
KL divergence & 0.0977 & 0.5838 \\
\hline
Probabilistic associator loss & 0.1453 & 0.8212 \\
\hline
\end{tabular}
\end{table*}
\begin{table*}
\caption{Comparison of impacts of training set choice on guess and associative rates (AL network).}
\label{table2}
\begin{tabular}{l|c|c|c|c}
\hline
Metric & min & average & max & std deviation \\
\hline
Guess rate & 0.1362 & 0.13979 & 0.1463 & 0.0036 \\
\hline
Associative rate & 0.7878 & 0.8181 & 0.8468 & 0.0187 \\
\hline
\end{tabular}
\end{table*}

\end{document}